\documentclass{article} 
\usepackage[preprint]{technical_template}

\usepackage{microtype}
\usepackage{hyperref}
\usepackage{url}
\usepackage{booktabs}

\usepackage{tcolorbox}
\tcbuselibrary{breakable}

\usepackage{lineno}
\usepackage{mymathdef}
\newcommand{\task}{{\textsc{CrossMath}}}

\definecolor{darkblue}{rgb}{0, 0, 0.5}
\hypersetup{colorlinks=true, citecolor=darkblue, linkcolor=darkblue, urlcolor=darkblue}

\title{Do Vision-Language Models Truly Perform Vision Reasoning? A Rigorous Study of the Modality Gap}

\author{Yige Xu${^{1,2,*}}$, Yongjie Wang$^{2,*}$, Zizhuo Wu$^1$, Kaisong Song$^3$, Jun Lin$^3$, Zhiqi Shen$^{1,\dagger}$\\
$^1$College of Computing and Data Science, Nanyang Technological University, Singapore \\
$^2$Alibaba-NTU Global e-Sustainability CorpLab (ANGEL) \\
$^3$Tongyi Lab, Alibaba Group, China\\
\texttt{yige002@e.ntu.edu.sg}, \texttt{\{yongjie.wang,zqshen\}@ntu.edu.sg}
}

\begin{document}

\iftechnicalreportsubmission
\linenumbers
\fi

\maketitle

\begin{abstract}
Reasoning in vision-language models (VLMs) has recently attracted significant attention due to its broad applicability across diverse downstream tasks. However, it remains unclear whether the superior performance of VLMs stems from genuine vision-grounded reasoning or relies predominantly on the reasoning capabilities of their textual backbones. To systematically measure this, we introduce {\task}, a novel multimodal reasoning benchmark designed for controlled cross-modal comparisons. Specifically, we construct each problem in text-only, image-only, and image+text formats guaranteeing identical task-relevant information, verified by human annotators. This rigorous alignment effectively isolates modality-specific reasoning differences while eliminating confounding factors such as information mismatch. Extensive evaluation of state-of-the-art VLMs reveals a consistent phenomenon: a substantial performance gap between textual and visual reasoning. Notably, VLMs excel with text-only inputs, whereas incorporating visual data (image+text) frequently degrades performance compared to the text-only baseline. These findings indicate that current VLMs conduct reasoning primarily in the textual space, with limited genuine reliance on visual evidence. To mitigate this limitation, we curate a {\task} training set for VLM fine-tuning. Empirical evaluations demonstrate that fine-tuning on this training set significantly boosts reasoning performance across all individual and joint modalities, while yielding robust gains on two general visual reasoning tasks. Source code is available at \url{https://github.com/xuyige/CrossMath}.
\begingroup\def\thefootnote{$*$}\footnotetext{The first two authors contributed equally.}\def\thefootnote{$\dagger$}\footnotetext{Corresponding authors.}\endgroup
\end{abstract}

\section{Introduction}
\label{sec:intro}
Building upon the profound success of Large Language Models (LLMs)~\cite{openai2023gpt4,dubey2024llama,DBLP:journals/corr/abs-2412-15115,deepseek2025deepseekr1,qwen3}, recent advancements have rapidly propelled the development of Vision-Language Models (VLMs)~\cite{DBLP:conf/nips/LiuLWL23a,qwen2026qwen35,singh2026openai}. By seamlessly integrating visual inputs with pure text, these models exhibit formidable potential in a diverse array of applications, ranging from image captioning and visual question answering to document understanding and visual grounding. To achieve this broad multimodal intelligence, modern VLMs typically rely on a standardized modular pipeline: a vision encoder extracts visual features, a cross-modal projector aligns these representations with the latent language space, and a pre-trained text decoder performs the final autoregressive generation~\cite{DBLP:conf/nips/LiuLWL23a,qwen2026qwen35}. Despite their impressive performance across multimodal benchmarks, it remains largely underexplored \textit{whether these models genuinely engage in visual reasoning, or merely exploit the inherent reasoning capabilities of their textual backbones.}

Disentangling genuine visual reasoning from textual reliance has thus emerged as a central problem in evaluating modern VLMs. However, existing benchmarks consistently fail to disentangle these modalities separately. On one hand, many existing baselines \cite{10.5555/3692070.3694451,yue-etal-2025-mmmu,DBLP:conf/cvpr/YueNZ0LZSJRSWYY24} either evaluate merely surface-level visual recognition or heavily exploit textual priors. They fail to satisfy the rigorous demands of visually intensive tasks that require multi-step spatial and geometric reasoning grounded entirely in the visual space. Consequently, these benchmarks fall short in capturing the nuanced differences in the genuine visual reasoning capabilities of VLMs. On the other hand, although newer benchmarks \cite{hao2025can,yao2025mmreason,stogiannidis2025mind,xu2026visulogic} introduce complex multimodal scenarios such as mathematics, physics, and chemistry problems, their problem formulations are often deeply entangled—requiring both visual and textual inputs simultaneously. Because the absence of either modality makes the question inherently unsolvable, these entangled tasks cannot be used to isolate and evaluate modality-specific reasoning capacities.

To rigorously analyze genuine visual reasoning ability, we argue that an effective evaluation must satisfy three core principles. First, \textit{tasks must be intrinsically ``vision-first.''} Achieving optimal performance should heavily depend on reasoning over spatial, geometric, or physical dynamics. In other words, the tasks must provide both step-by-step signals to verify the intermediate visual reasoning process, as well as definitive ground-truth answers to evaluate the correctness of the final output. Second, \textit{the dataset should encompass a stratified distribution of problem difficulties. }Systematically controlling the difficulty prevents performance saturation or floor effects, thereby allowing the benchmark to effectively differentiate the reasoning capacities of VLMs across varying parameter scales. Third, \textit{the benchmark must provide strictly equivalent questions across visual and textual formats.} This guarantees that any differences in performance stem entirely from the model's modality-specific reasoning capacities, rather than from incomplete information. By eliminating the confounding effects of information asymmetry, we ensure that the absence of either modality does not render the problem unsolvable.

Based on the aforementioned discussion, we introduce {\task}, a rigorously designed multimodal reasoning benchmark to quantitatively isolate and evaluate visual-textual reasoning capabilities. {\task} tasks the VLMs with inferring missing values within a 2D spatial grid of intersecting mathematical equations, outputting the predicted numbers sequentially (from top to bottom, left to right). This design explicitly satisfies our three evaluation principles: First, the 2D layout of intersecting equations intrinsically demands spatial geometric understanding and step-by-step logical deduction, providing clear intermediate signals and definitive ground-truth answers. Second, the procedural generation allows us to precisely control difficulty levels by adjusting grid sizes, the number of missing equations, and the complexity of operators, thereby guaranteeing sufficient discriminative power to evaluate VLMs across diverse parameter scales.
Finally, to eliminate modality confounding, each {\task} puzzle is formulated into three strictly equivalent formats—an image-only grid, a text-only markdown table, and an image+text prompt—ensuring identical task-relevant information across all settings.

To support rigorous evaluation and demonstrate the efficacy of post-training, we construct the {\task} benchmark, featuring three difficulty levels with $5,000$ training and $250$ evaluation samples. To ensure strict quality control, human annotators were recruited to manually verify the cross-modal information equivalence across all $250$ evaluation samples. Through extensive evaluations on state-of-the-art VLMs, we uncover a counterintuitive phenomenon: models achieve their highest performance with text-only inputs, experience unexpected degradation when visual data is integrated, and perform worst under vision-only conditions. This indicates that current VLMs rely predominantly on textual shortcuts rather than genuine visual reasoning. To mitigate this modality gap, we post-train Qwen3.5-9B on the {\task} training set using Supervised Fine-Tuning (SFT) and Group Relative Policy Optimization (GRPO) \cite{deepseek2025deepseekr1} with solely image-based inputs. Empirical results demonstrate that our post-training significantly boosts visual reasoning and effectively closes the performance gap across modalities. Furthermore, out-of-distribution evaluations show that this post-training preserves the model's original capabilities and yields consistent gains on external vision-based mathematical tasks.

The main contributions of this work are summarized as follows:

\noindent {\bf (1) Rigorous Evaluation \& Benchmark}: We proposal a systematic methodology to measure modality-specific reasoning capacity in VLMs. To support this, we construct {\task}, a strictly controlled, multimodal-equivalent dataset that provides step-wise visual annotations for fine-grained reasoning evaluation.

\noindent {\bf (2) Exposure of the Modality Gap}: Through systematic evaluation of state-of-the-art VLMs, we empirically demonstrate that these models predominantly rely on text-level reasoning shortcuts, often treating visual inputs as secondary and detrimental to performance.

\noindent {\bf (3) Effective Post-Training \& Robust Transfer}: We establish that image-only post-training is highly effective in rectifying these deficits, not only fostering genuine visual grounding but also driving robust out-of-distribution transfer without compromising the model's inherent capabilities.

\section{Related Works}

\subsection{Measuring the Visual-Textual Reasoning Gap in VLMs}

Although textual reasoning has been widely explored by the community~\cite{DBLP:conf/nips/Wei0SBIXCLZ22,DBLP:conf/nips/YaoYZS00N23,DBLP:conf/iclr/0002WSLCNCZ23,xu2025softcot,xu2025softcotpp}, a growing body of work suggests that strong language-side reasoning in Vision-Language Models (VLMs) does not automatically translate into visually grounded reasoning. Early studies connect failures in spatial reasoning to weak object localization and grounding, showing that perceptual imprecision can propagate into downstream reasoning errors~\cite{rajabi2023towards,DBLP:conf/icml/00020Z0GXLH25}. More recent benchmarks reinforce this limitation: state-of-the-art VLMs remain brittle on spatial reasoning, chart understanding, ARC-style transformations, and other settings in which success depends on visual structure rather than linguistic priors or knowledge recall~\cite{stogiannidis2025mind,unsal2025easyarc,tang2025chartmuseum,xu2026visulogic}. Related work on visualized text further shows that even semantically equivalent content can become substantially harder once it is rendered visually rather than provided as plain text, highlighting a persistent gap between language-space reasoning and image-grounded reasoning~\cite{liu2026vista}. Mechanistic analyses likewise suggest that perception and reasoning remain only weakly coupled in current VLMs~\cite{DBLP:conf/icml/00020Z0GXLH25,li2025unveiling}.

Despite these advances, existing studies do not yet provide a fully controlled measurement of modality-specific reasoning. Some benchmarks are diagnostic of visual failures, but do not offer strictly matched text-only and image-only versions of the same problem. Others evaluate multimodal reasoning in domains such as mathematics and science, but their tasks are inherently modality-entangled: the image and text are complementary rather than interchangeable, so removing either modality changes task solvability~\cite{DBLP:conf/cvpr/YueNZ0LZSJRSWYY24,yue-etal-2025-mmmu,DBLP:conf/eccv/ZhangJZLGQZLCQGL24,hao2025can,yao2025mmreason}. As a result, cross-modality performance differences are difficult to interpret, because they may reflect information asymmetry rather than modality-specific reasoning ability. {\task} is designed to address this gap by constructing semantically equivalent text-only, image-only, and image+text versions of the same vision-first puzzle, enabling direct comparisons of reasoning performance across modalities.

\subsection{Visual Reasoning Benchmarks}

Visual reasoning benchmarks span a broad family of tasks, including inductive, analogical, algorithmic, deductive, and spatial/geometric reasoning~\cite{lymperaiou2026reasoning}. Early abstract-puzzle benchmarks such as PuzzleVQA deliberately minimize dependence on world knowledge and instead emphasize rule induction over attributes such as number, color, shape, and size~\cite{chia-etal-2024-puzzlevqa}. More recent datasets extend this agenda through knowledge-light visual puzzles, grid-based reasoning tasks, and ARC-style transformations that require multi-step inference and self-correction~\cite{song2025visualpuzzles,ren2025vgrp,unsal2025easyarc}.

A complementary line of work focuses on concept-based and spatially grounded reasoning. Bongard-style datasets test whether models can infer latent concepts from sets of positive and negative visual examples~\cite{DBLP:conf/icml/WustTHISDRK25}, while spatial reasoning benchmarks probe relative position, layout understanding, planning, and inference over partially observed scenes in both abstract and natural-image settings~\cite{DBLP:conf/emnlp/MayerBJNB25,DBLP:conf/emnlp/LyuZYLJY25,pothiraj2025capture,khezresmaeilzadeh2026vriq}. Together, these benchmarks have shown that many VLMs struggle when reasoning depends on geometry, topology, or hidden structure rather than semantic priors.

Related multimodal math and science benchmarks, including MMMU/MMMU-Pro, MathVerse, EMMA, and MMReason, push models toward more realistic expert-level reasoning over diagrams, figures, and textual context~\cite{DBLP:conf/cvpr/YueNZ0LZSJRSWYY24,yue-etal-2025-mmmu,DBLP:conf/eccv/ZhangJZLGQZLCQGL24,hao2025can,yao2025mmreason}. These datasets are valuable for evaluating end-to-end multimodal competence, but they are not designed to isolate modality-specific reasoning because their visual and textual components are often jointly necessary. {\task} complements this literature by focusing on structured visual-symbolic reasoning under strict cross-modal equivalence: it is vision-first, supports step-wise supervision, spans multiple difficulty levels, and preserves task-relevant information across text-only, image-only, and image+text formulations.

\section{Preliminaries and Definitions}

\subsection{Visual Reasoning Measurements}

To evaluate the visual reasoning capabilities of vision-language models (VLMs), a natural approach is to measure their performance on reasoning benchmarks with visual inputs. However, results obtained solely under a vision-only setting are insufficient to determine whether observed limitations arise from the model’s underlying reasoning deficits or from the additional challenges of perceiving, encoding, and grounding visual information. In other words, poor performance in the visual modality may arise not solely from weak reasoning, but also from errors introduced during visual processing. To disentangle these factors, the same tasks should also be evaluated under text-only, image-only and hybrid-modality settings, where the reasoning requirements remain unchanged but the form of input varies. Such a setup makes it possible to quantify the modality gap, isolating the performance degradation strictly attributable to visual processing.

For these comparisons to be meaningful, the inputs across modalities must be semantically equivalent, so that performance differences can be attributed to modality rather than to confounding variations in task formulation, data format, or problem difficulty. Based on this consideration, a benchmark designed to measure the visual--textual reasoning gap in VLMs should satisfy the following three principles:

\noindent {\bf (1) The tasks should be inherently vision-first,} so that solving them genuinely requires visual understanding rather than relying primarily on textual shortcuts or external knowledge.

\noindent {\bf (2) The dataset should span multiple levels of difficulty,} allowing analysis of not only overall performance but also how the visual--textual gap evolves as reasoning complexity increases.

\noindent {\bf (3) The benchmark should provide strictly equivalent visual and textual formulations of the same questions,} ensuring that cross-modality comparisons are controlled, fair, and directly interpretable.

\subsection{{\task} Task Definition}

The input of the $j$-th example from {\task} includes a task-specific instruction $\cI=[i_1,i_2,\cdots,i_{|\cI|}]$ and an input query $\cQ_j$ that
\begin{align}
  \cQ_j =
  \begin{cases}
  \{\cQ_{\text{text},j}, \cQ_{\text{image},j}\}, & \text{if multi-modal} \\
  \{\cQ_{\text{text},j}\}, & \text{if textual-only} \\
  \{\cQ_{\text{image},j}\}, & \text{if vision-only}
  \end{cases}
\end{align}
where $\cQ_{\text{text},j}$ is equivalent to $\cQ_{\text{image},j}$ but in different modalities. Every $\cQ_j$ has multiple arithmetic reasoning problems, and the ground truth contains the answer of arithmetic reasoning problems from top to bottom, from left to right:
\begin{align}
  \cA_j=\mathrm{Ordered}(a_{1,j}, a_{2,j},\cdots,a_{|\cA_j|,j}).
\end{align}

Based on the input, we formulate the {\task} solving process of an VLM in two auto-regressive stages:

\noindent (1) \textbf{Structured Reasoning}: VLMs would produce explicit step-wise structured reasoning $\hat{\cS}_j$ based on the input:
\begin{align}
  \hat{\cS}_j&=[\hat{s}_{1,j},\hat{s}_{2,j},\cdots,\hat{s}_{|\hat{\cS}_j|}],\\\nonumber
  \hat{s}_{i,j}&=\mathrm{VLM}(\cI,\cQ_j,\hat{s}_{<i,j}),
\end{align}
where $\hat{s}_{i,j}=[t_{1,j}^{(i)}, t_{2,j}^{(i)},\cdots,t_{n_i,j}^{(i)}, \hat{a}_{i,j}]$ indicates the $i$-th intermediate step for structured reasoning, $t_{1,j}^{(i)}, t_{2,j}^{(i)},\cdots,t_{n_i,j}^{(i)}$ indicates the rationale tokens for the $i$-th step, and $\hat{a}_{i,j}$ indicates the predicted answer span for the $i$-th problem for the $j$-th example.

\noindent (2) \textbf{Final Answer Prediction}: With the answers from structured reasoning, VLMs would construct the final answer under the order from left to right, from top to bottom:
\begin{align}
  \hat{\cA}_j=\mathrm{Ordered}(\hat{a}_{1,j}, \hat{a}_{2,j}, \cdots, \hat{a}_{|\hat{\cS_j}|,j}).
\end{align}

After the final answer prediction, we evaluate the prediction with 6 fine-grained metrics (See \S~\ref{sec:exp-metrics} for metric details), which is formulated as:
\begin{align}
  \mathrm{Score}_j=f(\hat{\cA_j},\cA_j),
\end{align}
where $\cA_j$ is the sorted groud truth answer of the $j$-th example, and $f(\cdot)$ is the evaluation metric.

\section{Methodology}
\begin{figure*}[h!]
\centering
\subfloat[Textual Markdown Table]{
    \includegraphics[width=0.4\linewidth]{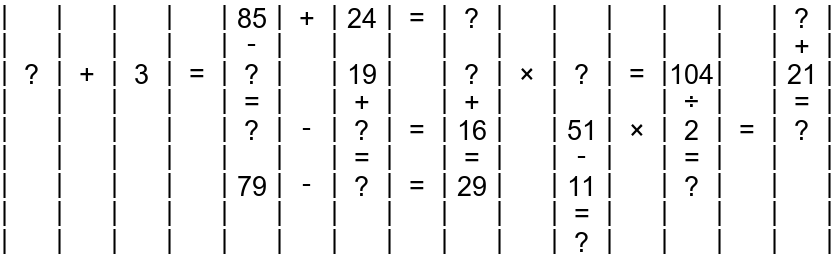}
    \label{fig:dataset-input-markdown}
}\\
\subfloat[Raw Image Input]{
    \includegraphics[width=0.4\linewidth]{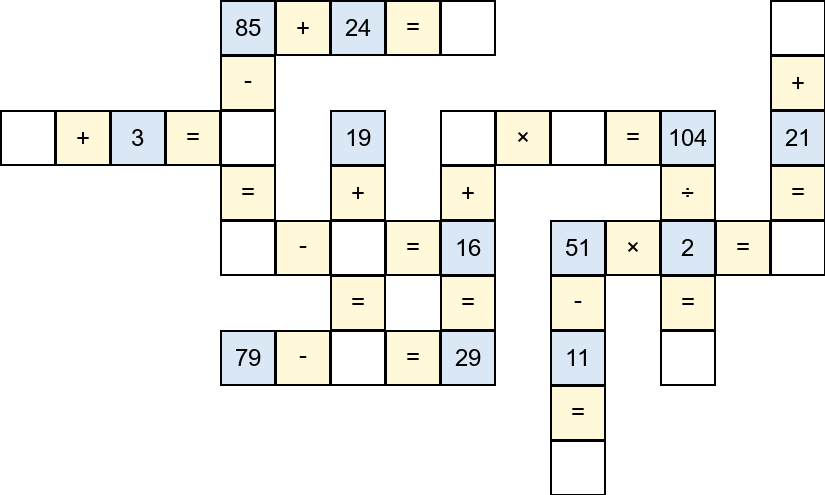}
    \label{fig:dataset-input-example}
}
\subfloat[Solutions]{
    \includegraphics[width=0.4\linewidth]{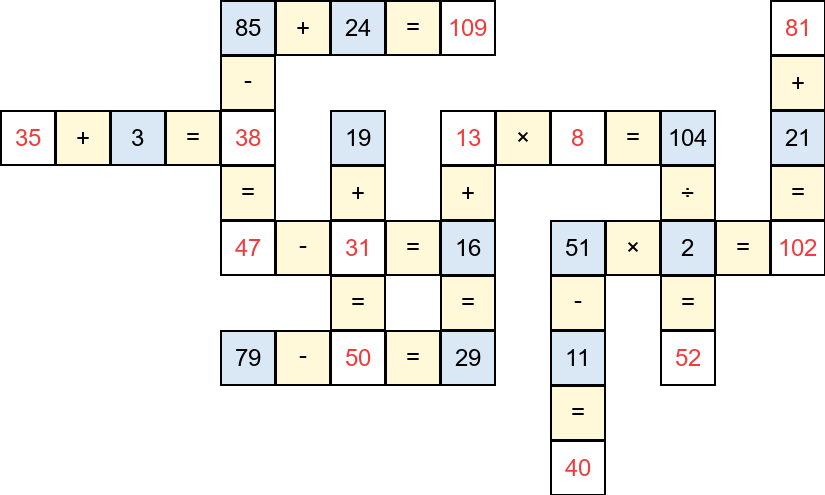}
    \label{fig:dataset-output-example}
}
\caption{Example images of our {\task} puzzle. Target solutions are marked in red. \label{fig:dataset-example}}
\end{figure*}

\begin{figure*}[h!]
\centering
\subfloat[Original Style]{
    \includegraphics[width=0.3\linewidth]{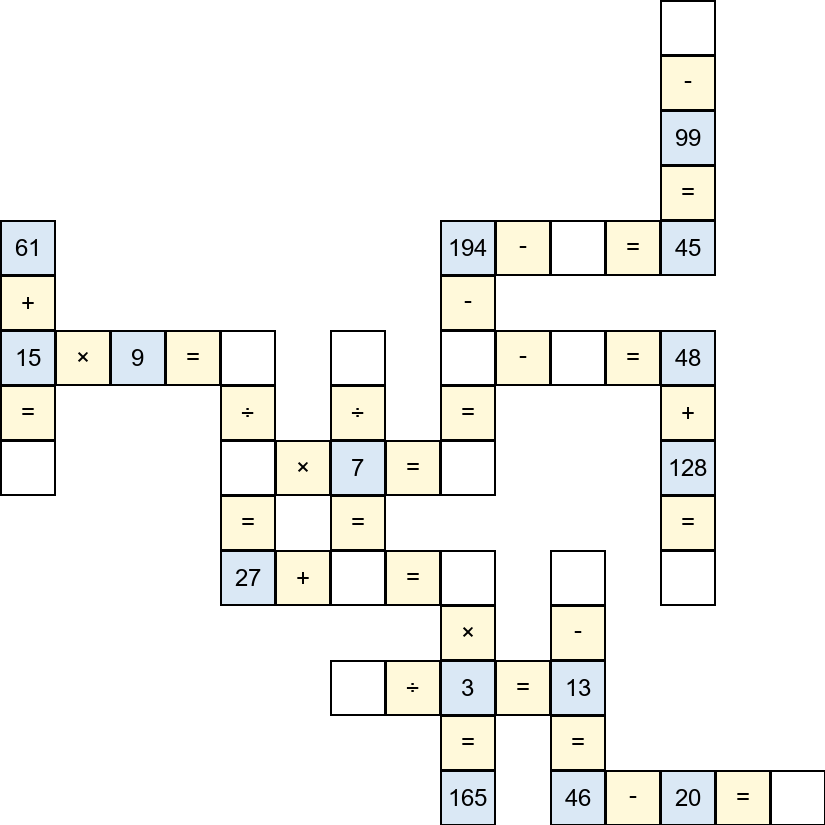}
    \label{fig:dataset-ori-input-example}
}
\subfloat[Without Border]{
    \includegraphics[width=0.3\linewidth]{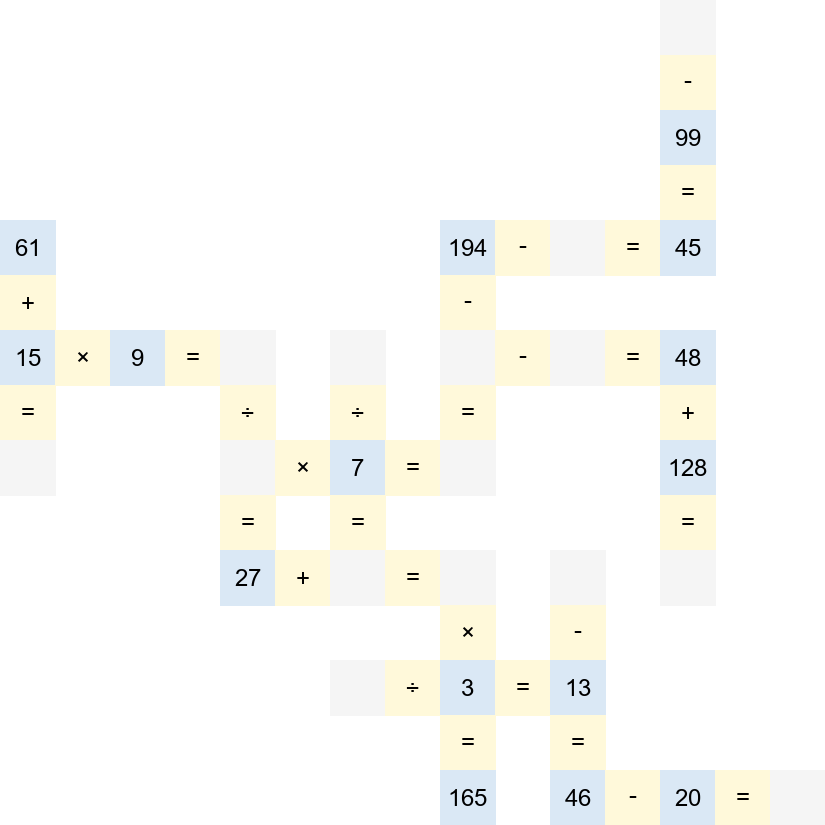}
    \label{fig:dataset-noboarder-input-example}
}\\
\subfloat[With Significant Background]{
    \includegraphics[width=0.3\linewidth]{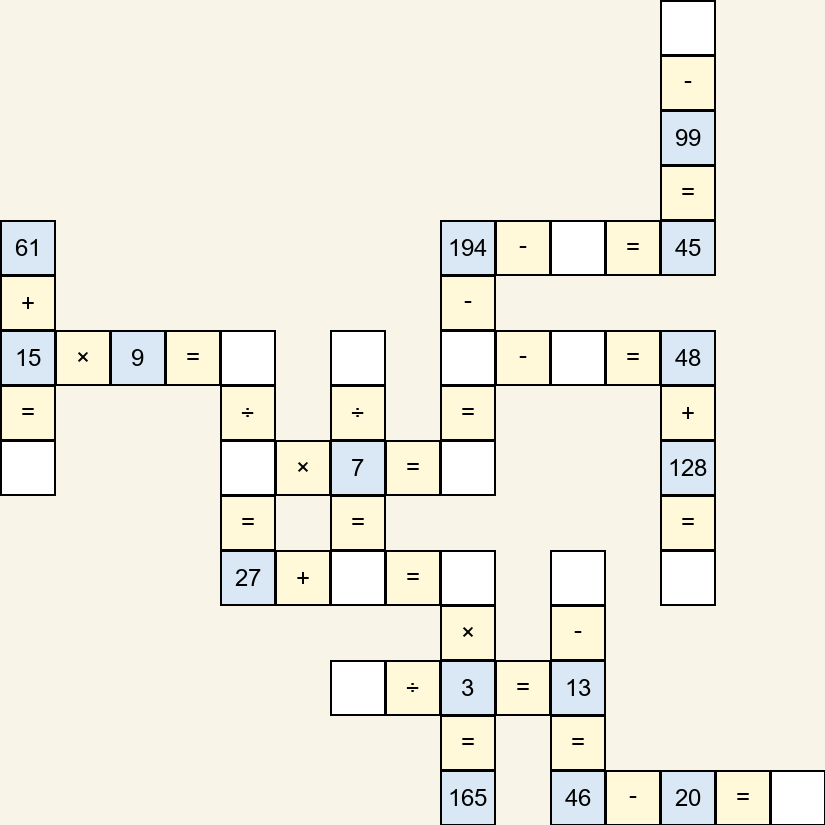}
    \label{fig:dataset-beige-input-example}
}
\subfloat[Change Font and Color]{
    \includegraphics[width=0.3\linewidth]{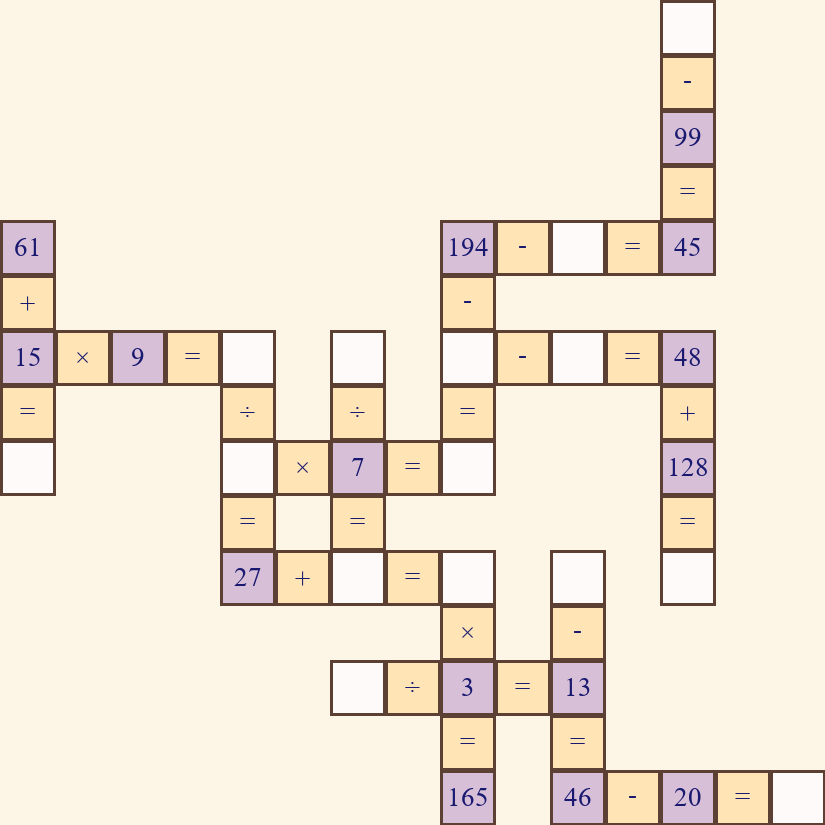}
    \label{fig:dataset-altstyle-input-example}
}
\caption{Example images for different vision styles or formats of our {\task} puzzle.
\label{fig:dataset-test-example}}
\end{figure*}

\subsection{Dataset Curation}
To construct the {\task} benchmark, we developed an automated data curation pipeline comprising four key stages: (1) Automated Collection, where raw cross-math puzzles in image format and their corresponding solutions are scraped from web sources; (2) Multimodal Alignment, which utilizes text recognition techniques to convert image-based puzzles into structured Markdown format; (3) Reasoning Path Extraction, which identifies and formalizes intermediate step-by-step solutions; and (4) Style Augmentation, which generates diverse visual styles for the puzzle images to enhance dataset variety. Figure~\ref{fig:dataset-example} illustrates a representative {\task} instances generated by this pipeline.

\subsubsection{Raw Data Collection} In this stage, we leverage Playwright~\footnote{https://github.com/microsoft/playwright} to programmatically interface with an online arithmetic puzzle generation website ~\footnote{https://www.ohmydots.com/creator-cross-math.html}. To control diversity in the collected samples, we systematically vary the generation parameters, including difficulty levels (Easy, Medium, or Hard), operator sets (random combinations of $\{+, -, \times, \div\}$), numeric ranges (e.g., 50--250), and the total number of equations (e.g., 5--15). For each generated instance, we capture two high-resolution screenshots: an unsolved query version and a solution-annotated version, as shown in Figure~\ref{fig:dataset-input-example} and Figure~\ref{fig:dataset-output-example}. Furthermore, by parsing the webpage HTML data, we extract the underlying textual equations and answers, providing a rigorous source for ground-truth verification.

\subsubsection{Image to Markdown Table} In this stage, we transform each visual puzzle into a structured Markdown representation using specialized image processing heuristics. Specifically, we leverage pixel-level color differences to categorize the functional role of each grid cell: blue cells represent fixed constants, white cells with red text indicate unknown variables (targets), and yellow cells denote arithmetic operators. To facilitate efficient and accurate recognition, individual cells are cropped and arranged into an indexed composite image (i.e., a tiled mosaic) with explicit markers (e.g., [1], [2], \ldots). This consolidated image is then processed by Qwen3-VL-Max~\cite{qwen3} for high-precision batch OCR, enabling the simultaneous extraction of all numerical and symbolic content while maintaining spatial correspondence. By integrating the recognized cell contents with their original spatial coordinates, we reconstruct each puzzle into a structured Markdown table.

To mitigate common OCR errors such as the confusion of the digit `1' with visually similar characters like `l' or `$|$', we implement rule-based post-processing, including regex-based heuristics and syntax-aware corrections. Following this automated step, we perform a manual audit of all processed samples to ensure absolute data fidelity.  This rigorous workflow ensures strict information equivalence between the visual and textual modalities, eliminating any potential modality bias resulting from information asymmetry.

\subsubsection{Reasoning Path Extraction} To derive the ground-truth reasoning chain, we implement a symbolic solver that transforms the 2D grid into an ordered sequence of logical deductions. We first formalize the incomplete puzzle and its corresponding solution into two-dimensional arrays, denoted as $\mathcal{G}_Q$ and $\mathcal{G}_A$, respectively.

\textbf{Equation Topology Detection.} We traverse $\mathcal{G}_Q$ to identify valid equation structures across both horizontal and vertical axes. A horizontal equation is detected if a numeric cell is followed by an operator at $(x, y+1)$ and an equality sign at $(x, y+3)$. Similarly, a vertical equation is identified if these elements appear at $(x+1, y)$ and $(x+3, y)$, respectively. For each detected topology, we record the coordinates of the three participating numeric cells and retrieve their ground-truth values from $\mathcal{G}_A$. These values are then aligned with the textual metadata extracted during Stage 1 to reconstruct the complete symbolic expression.

\textbf{Iterative Symbolic Deduction.} To simulate a human-like solving process, we initialize a set of $known\_cells$, seeded with the numbers provided in the original puzzle. We then perform iterative deduction to resolve the unknown variables. In each iteration, the solver examines all unresolved equations: if at least two of the three numeric quantities in an equation are contained within $known\_cells$, the equation is flagged as `solvable.'

All equations resolved within the same iteration are grouped into a single reasoning step. Crucially, newly resolved coordinates are merged into $known\_cells$ only after the current iteration is finalized. This synchronized update mechanism ensures that each step depends exclusively on information established in previous iterations, thereby preserving a strict causal structure for multi-step reasoning. Formally, each step $s_i$ is represented as:
\begin{equation*}
s_i = { (eq_{j}, ans_{j}) \mid eq_{j} \text{ is solvable at iteration } i }
\end{equation*}
where $eq_{j}$ denotes the $j$-th equation and $ans_{j}$ is its corresponding numerical solution. This process terminates once all equations are resolved, yielding a chronologically ordered dictionary of solving steps that serves as the gold-standard Chain-of-Thought (CoT) for each {\task} instance.

\subsubsection{Visual Style Augmentation}
To evaluate the visual robustness of VLMs and ensure they generalize beyond specific rendering artifacts, we implement a Style Augmentation stage. As illustrated in Figure~\ref{fig:dataset-test-example}, we systematically perturb the original puzzle images through four distinct visual transformations:
\begin{itemize}
\item Original Style: The default rendering from the source website, featuring standard grid lines and color-coded cells.
\item Border Removal: We eliminate the bounding boxes and grid lines for each cell, forcing the model to rely solely on the relative spatial positioning of digits and operators.
\item Background Complexity: We introduce non-uniform background textures or significant color fills to test the model's ability to distinguish foreground content from visual noise.
\item Font and Palette Variation: We stochastically vary the font styles, and color schemes (e.g., swapping background and text colors) to prevent the model from exploiting low-level color priors.
\end{itemize}
By generating these diverse visual formats, we ensure that {\task} serves as a rigorous testbed for evaluating genuine visual reasoning rather than pattern matching on fixed image templates.

The statistics of the benchmark
dataset are summarized in Table~\ref{tab:testset_stats}. The benchmark comprises 250 core {\task} instances, each featuring a raw puzzle image paired with its corresponding ground-truth Markdown table to ensure information symmetry. By applying the four visual augmentation styles described above, we expand these core instances into a total of 1,000 image-based evaluation samples. This diverse test suite allows us to benchmark VLMs across three distinct input configurations: image-only, text-only, and multimodal (image + text).

\subsection{Post-Training}
Our preliminary study demonstrates that modern VLMs exhibit significant performance degradation in image-only settings. In this section, we investigate whether a post-training strategy with image only can compensate for these visual reasoning inefficiencies and bridge the observed modality gap. To achieve it, we compile a training dataset comprising $5,000$ unique {\task} puzzle images and their corresponding solutions. To bolster the model's visual robustness, we systematically pair each original puzzle with its background-augmented variant. Notably, all training instances and test samples underwent a cross-comparison and manual check, ensuring that the two sets are strictly disjoint.

\subsubsection{Supervised Fine-Tuning (SFT)}
Observing that small-scale VLMs often fall into redundant and vacillating reasoning loops, even exhausting the context window without yielding an answer, we first conduct SFT to internalize structured thinking patterns.

To provide high-quality supervision, we construct reasoning trajectories by prompting Qwen3-VL-Max \cite{qwen3} to verbalize the symbolic steps into coherent CoT sequences given the Markdown tables. While Qwen3-VL-Max occasionally produce wrong answers, we intentionally utilize the all generated reasoning chain to facilitate a behavioral cold-start. This strategy is designed to instill the logical syntax and procedural sequence of multi-step reasoning within the student VLM.  We fine-tune the Qwen3.5-9B \cite{qwen2026qwen35} model via SFT on our image-reasoning corpus, utilizing LoRA \cite{DBLP:conf/iclr/HuSWALWWC22} for efficient adaptation. More implementation details for SFT is shown in \S~\ref{sec:implementation_details_sft}.

\subsubsection{Reinforcement Learning with Verifiable Rewards (RLVR)} Recently, RL has gained significant traction as a powerful paradigm \cite{rafailov2023direct,meng2024simpo,deepseek2025deepseekr1} for enhancing LLM performance across a broad spectrum of reasoning-intensive tasks, including mathematical problem-solving, code generation, and complex agentic interactions. Unlike conventional actor-critic frameworks \cite{rafailov2023direct,meng2024simpo}, Group Relative Policy Optimization (GRPO) \cite{deepseek2025deepseekr1} obviates the need for a standalone critic model by employing a group-based preference mechanism. By computing the advantage based on the relative performance within a group of outputs, GRPO provides a more scalable and efficient optimization paradigm for LLM fine-tuning.
While rule-based rewards are common in GRPO, extracting intermediate results from raw reasoning chains can be brittle. We instead introduce a position-weighted reward strategy that implicitly supervises the reasoning process through its outcomes. By assigning larger weights to sub-problems requiring more reasoning hops and lower weights to preliminary sub-problems, we incentivize the model to solve the entire logical sequence. The final reward $r$ is calculated as the weighted accuracy across all target cells:
\begin{equation}r_j = \frac{\sum_{i=1}^{|\mathcal{S}_j|} w_i\cdot \mathbb{I}[\hat{a}_{i,j} = a_{i,j}]}{\sum_{i=1}^{|\mathcal{S}_j|} w_i},\end{equation}
where $\mathbb{I}[\cdot]$ is the indicator function, $\hat{a}_{i,j}$ and $a_{i,j}$ are predicted answers and the ground-truth answers for the $i$-th problem in the $j$-th example, respectively, and $w_i$ scales with the logical depth (hops) of the problem. More implementation details for RLVR is shown in \S~\ref{sec:implementation_details_rlvr}.

\begin{table*}[tb]
\centering
\begin{tabular}{lcccccc}
\toprule
\textbf{Difficulty} & \textbf{\# Examples/\%} & \textbf{Avg. Problems} & \textbf{1 Hop} & \textbf{2 Hops} & \textbf{3 Hops} & \textbf{4+ Hops} \\
\midrule
Easy & 90 / 36.0\% & 10.46 & 10.46  & 0.00  & 0.00  & 0.00 \\
Medium & 85 / 34.0\% & 9.81 & 7.66  & 1.53 & 0.53  & 0.09 \\
Hard & 75 / 30.0\% & 10.64 & 5.13 & 2.52 & 1.73 & 1.25 \\
\midrule
Total & 250  & 10.29 & 7.91 & 1.28  & 0.70 & 0.41  \\
\bottomrule
\end{tabular}
\caption{Statistics of the {\task} test set across different difficulty levels and reasoning hop levels (total $N=250$). Statistics of the {\task} test set stratified by different difficulty levels and reasoning hop levels. Each example associates with several problems to be filled; we report average number of to address, the problems across reasoning depths requires 1 hop, 2 hops, 3 hops, and 4+ hops reasoning are also reported. }
\label{tab:testset_stats}
\end{table*}

\section{Experiments}

\subsection{Metrics}
\label{sec:exp-metrics}

In {\task}, we consider the following evaluation metrics:

\noindent {\bf (1) Micro Accuracy} considers the partial correctness of the reasoning chains that every single correct problem solution would be counted:
\begin{align}
  \mathrm{Micro}_{\mathrm{acc}}(\hat{\cA},\cA)=\frac{1}{N}\sum_{j=1}^{N}\Big(\frac{1}{|\cA_j|}\sum_{n=1}^{|\cA_j|}\mathbb{I}(\hat{a}_{n,j}=a_{n,j})\Big),
\end{align}
where $\mathbb{I}(\cdot)$ is an indicator function.

\noindent {\bf (2) Macro Accuracy} considers the strict correctness of the whole reasoning chains that only all problems in an example are correct would be counted:
\begin{align}
  \mathrm{Macro}_{\mathrm{acc}}(\hat{\cA},\cA)=\frac{1}{N}\sum_{j=1}^{N}\Big(\mathbb{I}(\hat{\cA}_{j}=\cA_{j})\Big),
\end{align}

\noindent {\bf (3) $K$-Hop Accuracy} considers the accuracy of the problems that only requires a $K$-hop reasoning:
\begin{align}
  K\text{-Hop}_{\mathrm{acc}}(\hat{\cA},\cA)&=\frac{1}{|\bar{\cA}_K|}\sum_{a_{n,j}\in\bar{\cA}_K}\mathbb{I}(\hat{a}_{n,j}=a_{n,j}),\\\nonumber
  \bar{\cA}_K&=\{a_{n,j}|a_{n,j}\;\text{is the answer of}\;K\text{-Hop problem}\}
\end{align}

\subsection{Experiment Settings and Implementation Details}
To provide a thorough comparison, we benchmark our results against several state-of-the-art VLMs. This includes both open-source models and  proprietary models such as the Qwen3 \cite{qwen3} and Qwen3.5 \cite{qwen2026qwen35} families. These models support both unimodal (text-only or image-only) and multimodal input configurations. For a fair comparison, the same zero-shot CoT prompt is used regardless of the input format; All inference parameters, including temperature and sampling strategies, adhere to the default configurations specified in their model cards.

\subsubsection{Implementation Details for Supervised Fine-Tuning}
\label{sec:implementation_details_sft}

To obtain high-quality supervision signals, we construct reasoning trajectories by prompting Qwen3-VL-Max~\cite{qwen3} to verbalize symbolic reasoning steps into coherent chain-of-thought (CoT) sequences, conditioned on text-only queries paired with Markdown tables. The prompt \ref{prompt:trajectory} includes an example input used for Chain-of-Thought (CoT) trajectory generation.  To ensure brevity while preserving essential reasoning, we introduce a modified instruction compared to standard text-only queries: ``You are required to reason as **concisely** as you can, while keeping compulsory reasoning steps.'' In average, the generated CoT trajectories has around 5,200 tokens.

We implement all SFT experiments using the HuggingFace Transformers framework~\cite{wolf-etal-2020-transformers}. All models are trained on a single NVIDIA A100 80GB GPU. We adopt Low-Rank Adaptation (LoRA)~\cite{DBLP:conf/iclr/HuSWALWWC22} for parameter-efficient fine-tuning, with a rank of $r=16$, and train the model for 2 epochs. Optimization is performed using the AdamW optimizer~\cite{DBLP:conf/iclr/LoshchilovH19}, with a learning rate of $2\times10^{-5}$ and a weight decay of $0.01$.

We further employ a cosine learning rate schedule with a warmup ratio of $0.03$. Gradient accumulation is set to 8 steps to achieve an appropriate effective batch size. To avoid out-of-memory issues, we limit the maximum sequence length to 5,000 tokens during the supervised fine-tuning stage.

\subsubsection{Implementation Details for Reinforcement Learning with Verifiable Rewards}
\label{sec:implementation_details_rlvr}

Starting from a supervised fine-tuning (SFT) initialization, the VLM is capable of generating {\task} reasoning trajectories. Building upon this, we further apply reinforcement learning with verifiable rewards (RLVR) under the Group Relative Policy Optimization (GRPO) framework. The input format for RLVR remains identical to that used in SFT. For each instance, we sample 4 rollout sequences.

All RL experiments are implemented using the TRL framework~\cite{vonwerra2020trl}. Consistent with the SFT stage, we adopt Low-Rank Adaptation (LoRA)~\cite{DBLP:conf/iclr/HuSWALWWC22} for parameter-efficient fine-tuning, with a rank of $r=16$. The learning rate is set to $1\times10^{-6}$ to ensure stable training.

We set the maximum completion length to 6,000 tokens; rollouts exceeding this limit are treated as failures. In total, we perform 200 RL training steps.

\subsection{Main Results}

\begin{table*}
  \tabcolsep 2.0pt
  \centering
  \begin{tabular}{l|cc|cc|ccc}
    \toprule
     \multirow{2}{*}{Model} & \multicolumn{2}{c|}{Image Only} & \multicolumn{2}{c|}{Image + Text} & \multicolumn{2}{c}{Text Only} \\
    \cline{2-7}
    ~ & Micro Acc& Macro Acc & Micro Acc& Macro Acc & Micro Acc& Macro Acc \\
    \midrule
    \multicolumn{7}{c}{\it Zero-Shot}\\  \midrule
    Qwen3.6-Plus & 32.23 & 11.60 & {\bf 90.76} & {\bf 79.60} & 96.23 & 88.20 \\
    Qwen3.5-Plus & 35.65 & 12.40 & 85.22 & 74.80 & \textbf{97.27} & \textbf{92.80} \\
    Qwen3.5-397B-A17B & \underline{39.26} & {\bf 16.00} & \underline{87.54} & \underline{78.40} & \underline{96.86} & \underline{92.00} \\
    Qwen3.5-122B-A10B & 36.85 & \underline{12.80} & 81.37 & 61.20 & 89.61 & 77.60 \\
    Qwen3.5-27B & {\bf 40.74} & 12.40 & 87.04 & 66.40 & 88.67 & 75.20 \\
    Qwen3.5-9B & 23.25 & 3.20 & 61.56 & 29.60 & 73.39 & 44.00 \\
    \midrule

    \multicolumn{7}{c} {\it Supervised Training}\\ \midrule
    Qwen3.5-9B-SFT & \underline{59.52} & \underline{48.50} & 67.90 & 60.00  & 82.58 & 69.60 \\
    Qwen3.5-9B-SFT+GRPO & \textbf{62.33} & \textbf{50.40} & 71.21 & 62.80 & 87.36 & \underline{76.40} \\
  \bottomrule
\end{tabular}
\caption{VLM performance on {\task} under different modalities. The best results are indicated in bold, while the second-best are underlined.}\label{tab:exp-main-multimodal}
\end{table*}

In this study, we benchmark the performance of state-of-the-art VLMs across diverse input modalities and evaluate their step-wise reasoning capabilities. The detailed results are summarized in Tables~\ref{tab:exp-main-multimodal}, \ref{tab:exp-main-image}, and \ref{tab:exp-main-image1}. Based on these empirical findings, we derive several key observations.

\noindent {\bf(1) VLMs exhibit a significant reasoning performance gap between image and text modalities.} Despite the information symmetry across modalities, VLMs exhibit a stark performance degradation when reasoning with visual inputs compared to textual representations. For instance, while Qwen3.5-Plus achieves a Macro Accuracy of 92.8\% in the text-only setting, its performance drops to a mere 12.4\% when presented with the identical puzzle in an image format. This pervasive performance gap is observed across all evaluated models—ranging from Qwen3.5-9B and 27B to the flagship Qwen3.5-Plus. This consistency underscores a fundamental bottleneck: the models' current inability to reliably map visual evidence into symbolic representations that are essential for complex logical reasoning.

Ideally, the integration of visual information should complement textual reasoning, leading to performance that matches or exceeds text-only baselines. However, we observe a contrary trend: the introduction of visual inputs consistently leads to performance degradation. For instance, Qwen3.5-Plus drops from $97.27\%/92.80\%$ in the text-only setting to $85.22\%/74.80\%$ in the multimodal setting in Table \ref{tab:exp-main-multimodal}. This suggests that when equivalent textual information is already available, redundant visual inputs often fail to help and can reduce performance. Rather than aiding the process, visual evidence may interfere with the model’s internal logic by injecting ambiguous or poorly grounded features.

\noindent {\bf (2) Visual reasoning failures do not stem primarily from perception errors.} We find that perception errors are not the dominant cause of poor performance in the vision-only setting. On the one hand, when VLMs are tasked with converting puzzle images into Markdown tables, the OCR error rates remain remarkably low, suggesting that the models can accurately perceive the numerical content. On the other hand, if perception were the dominant bottleneck, providing reasoning-chain supervision would be futile, as no amount of logical optimization can compensate for missing or corrupted input data. The fact that our SFT and GRPO variants achieve marked improvements in the image-only modality suggests that the 'raw' visual features are present and accessible. The primary challenge is that VLMs do not lack the sight to recognize symbols, but rather the structural inductive bias to organize those symbols into a functional reasoning sequence. Finally, perceptual errors cannot explain the sharp performance decay associated with increasing reasoning hops. If perception were at fault, the difficulty would be independent of the number of logical steps. However, our data shows a precipitous decline in accuracy as reasoning depth scales, confirming that the dominant challenge in \textsc{CrossMath} is maintaining logical consistency across extended dependency chains, not the initial recognition of visual elements.

\noindent {\bf (3) Reasoning depth is the core bottleneck for both text and visual reasoning.} The discrepancy between Macro and Micro accuracy in the zero-shot setting reveals a critical insight: models can partially capture local reasoning patterns but frequently fail to maintain global logical correctness. For instance, in the text-only setting, Qwen3.5-9B achieves a Micro Accuracy of $73.39\%$ but only $44.00\%$ Macro Accuracy. A similar, albeit more pronounced, disparity is observed in the vision-only modality for Qwen3.5-Plus, which records a Micro Accuracy of $35.65\%$ and a Macro Accuracy of $12.40\%$.

From another perspective, the scaling of reasoning depth provides definitive evidence for the reasoning-centric nature of this bottleneck. Across all experimental settings regardless of image or text inputs, accuracy inversely correlates with the required number of logical hops. Specifically, Qwen3.5-Plus sees its performance fall by nearly 30\% when moving from 1-hop to 4+ hop problems. Similarly, the performance of the larger Qwen3.5-27B variant exhibits a precipitous drop from $42.26\%$ to $5.88\%$ in the vision-only setting. This trend highlights that while visual elements remain constant, the increasing demand for symbolic manipulation overwhelms the model. It confirms that the major difficulty is not recognizing symbols, but propagating logic across extended dependency chains, where any single failure in intermediate consistency leads to a total collapse of the reasoning performance.

\subsection{The Effectiveness of Post-training}
To evaluate the efficacy of our post-training pipeline in enhancing visual reasoning, we report the performance of our SFT and RL-tuned variants in Tables~\ref{tab:exp-main-multimodal} and \ref{tab:exp-main-image}. We observe the following phenomenon.

\noindent \textbf{(1) In-domain post-training serves as a catalyst for visual reasoning performance.} Supervised fine-tuning (SFT) on in-domain data leads to substantial improvements across all evaluation metrics. Notably, Qwen3.5-9B-SFT elevates Micro Accuracy from 23.25\% to 59.52\% and Macro Accuracy from 3.20\% to 48.50\%, demonstrating that task-specific supervision is highly effective in addressing logical reasoning deficiencies. Furthermore, the application of GRPO yields additional gains, particularly in multi-hop scenarios, suggesting that optimization at the reasoning trajectory level further refines the model's compositional reasoning ability. These results indicate that the failures observed in zero-shot settings are not necessarily inherent to the model architecture, but rather stem from a lack of appropriate training signals.

\noindent {\bf (2) Supervised training reduces but does not fully close the modality gap}: In-domain supervised training leads to substantial gains in the vision-only setting, showing that the modality gap can be alleviated through task-specific learning. Compared with zero-shot Qwen3.5-9B, both Qwen3.5-9B-SFT and Qwen3.5-9B-SFT+GRPO improve markedly on vision-only performance, indicating that the model can learn better visual parsing and task-specific reasoning strategies from aligned supervision. However, the text-only setting still remains clearly stronger after training. Even after SFT+GRPO, the model reaches 62.33\%/50.40\% in vision-only accuracy, which is still notably lower than its 87.36/76.40 text-only performance. This suggests that the remaining limitation is not merely a lack of in-domain examples, but may reflect a more fundamental architectural bottleneck in how visual evidence is encoded, projected, and integrated into the reasoning process. Overall, these results suggest that merely aligning visual embeddings with textual space is insufficient for complex logical tasks. Closing the modality gap requires more than just better projection layers; it necessitates more powerful visual foundation models capable of moving beyond pixel-level recognition to fully internalize the structural and physical constraints of the world.

\begin{table*}
  \centering
  \tabcolsep 2.0pt
  \begin{tabular}{l|cccccccccc}
    \toprule
    Model & Micro Acc& Macro Acc & 1 Hop Acc & 2 Hops Acc & 3 Hops Acc & 4+ Hops Acc \\
    \midrule
    \multicolumn{7}{c}{\it Zero-Shot}  \\ \midrule
    Qwen3.6-Plus & 32.23 & 11.60 & 31.08 & {\bf 16.84} & \underline{10.35} & 3.14 \\
    Qwen3.5-Plus & 35.65 & 12.40 & 37.13 & \underline{16.16} & 9.30 & \underline{8.82} \\
    Qwen3.5-397B-A17B & \underline{39.26} & {\bf 16.00} & \underline{39.86} & 15.15 & {\bf 11.05} & {\bf 9.80} \\
    Qwen3.5-122B-A10B & 36.85 & \underline{12.80} & 37.65 & 14.48 & 8.72 & 1.96 \\
    Qwen3.5-27B & {\bf 40.74} & 12.40 & {\bf 42.26} & 15.49 & 7.56 & 5.88 \\
    Qwen3.5-9B & 23.25 & 3.20 & 26.27 & 3.37 & 2.32 & 1.96 \\
    \midrule

   \multicolumn{7}{c} {\it Supervised Training}\\    \midrule
    Qwen3.5-9B-SFT & \underline{59.52} & \underline{48.50} & \underline{57.92} & \underline{53.96} & \underline{52.18} & \underline{40.93} \\
    Qwen3.5-9B-SFT+GRPO & \textbf{62.33} & \textbf{50.40} & \textbf{61.02} & \textbf{58.42} & \textbf{56.98} & \textbf{41.18} \\
  \bottomrule
\end{tabular}
\caption{VLM performance on {\task} (Image Only). The best results are indicated in bold, while the second-best are underlined.}
\label{tab:exp-main-image}
\end{table*}

\begin{table*}[t!]
  \centering
  \tabcolsep 4.0pt
  \begin{tabular}{l|cc|cccccccc}
    \toprule
    Model & Micro Acc& Macro Acc & 1 Hop Acc & 2 Hops Acc & 3 Hops Acc & 4+ Hops Acc \\
    \midrule
    \multicolumn{7}{c} {Text Only}\\ \midrule
    Qwen3.6-Plus & \underline{96.23} & \underline{88.20} & \underline{96.10} & \underline{93.94} & \underline{92.21} & \underline{88.63} \\
    Qwen3.5-397B-A17B & {\bf 96.86} & {\bf 92.00} & {\bf 97.07} & {\bf 95.62} & {\bf 93.02} & {\bf 93.14} \\
    Qwen3.5-122B-A10B & 89.61 & 77.60 & 88.30 & 83.84 & 80.23 & 74.51 \\
    Qwen3.5-27B & 88.67 & 75.20 & 87.45 & 83.84 & 80.23 & 81.37 \\
    Qwen3.5-9B & 73.39 & 44.00 & 73.93 & 56.90 & 48.26 & 29.41 \\
    Qwen3.5-9B-SFT & 82.58 & 69.60 & 83.16 & 82.83 & 76.74 & 65.69 \\
    Qwen3.5-9B-SFT+GRPO & 87.36 & 76.40 & 87.45 & 86.20 & 86.63 & 75.49 \\
    \midrule
   \multicolumn{7}{c}  {Image Only (Original style)}\\ \midrule
    Qwen3.6-Plus & 32.23 & 11.60 & 31.08 & 16.84 & 10.35 & 3.14 \\
    Qwen3.5-397B-A17B & 39.26 & 16.00 & 39.86 & 15.15 & 11.05 & 9.80 \\
    Qwen3.5-122B-A10B & 36.85 & 12.80 & 37.65 & 14.48 & 8.72 & 1.96 \\
    Qwen3.5-27B & 40.74 & 12.40 & 42.26 & 15.49 & 7.56 & 5.88 \\
    Qwen3.5-9B & 23.25 & 3.20 & 26.27 & 3.37 & 2.32 & 1.96 \\
    Qwen3.5-9B-SFT & \underline{59.52} & \underline{48.50} & \underline{57.92} & \underline{53.96} & \underline{52.18} & \underline{40.93} \\
    Qwen3.5-9B-SFT+GRPO & \textbf{62.33} & \textbf{50.40} & \textbf{61.02} & \textbf{58.42} & \textbf{56.98} & \textbf{41.18} \\
    \midrule
    \multicolumn{7}{c}  { Image Only (Without Border)}\\ \midrule
    Qwen3.6-Plus & 21.46 & 6.60 & 21.70 & 8.48 & 5.47 & 2.75 \\
    Qwen3.5-9B & 20.60 & 2.80 & 23.34 & 4.38 & 0.58 & 0.00 \\
    Qwen3.5-9B-SFT & \underline{55.87} & \underline{43.80} & \textbf{54.29} & \underline{46.63} & \underline{40.41} & \underline{35.78} \\
    Qwen3.5-9B-SFT+GRPO & \textbf{56.40} & \textbf{45.60} & \underline{54.16} & \textbf{47.92} & \textbf{44.19} & \textbf{36.27} \\
    \midrule
     \multicolumn{7}{c}  { Image Only (With Significant Background)}\\ \midrule
    Qwen3.6-Plus & 35.01 & 12.80 & 34.08 & 17.58 & 11.05 & 4.31 \\
    Qwen3.5-9B & 26.36 & 3.20 & 29.13 & 4.71 & 3.49 & 0.00 \\
    Qwen3.5-9B-SFT & \underline{59.87} & \textbf{50.40} & \underline{59.04} & \textbf{47.47} & \underline{41.28} & \underline{32.35} \\
    Qwen3.5-9B-SFT+GRPO & \textbf{60.19} & \underline{50.10} & \textbf{59.36} & \underline{46.63} & \textbf{44.19} & \textbf{36.27} \\
    \midrule
     \multicolumn{7}{c}  { Image Only (Change Font and Color)}\\ \midrule
    Qwen3.6-Plus & 36.30 & 14.70 & 34.73 & 19.66 & 13.49 & 5.10 \\
    Qwen3.5-9B & 25.66 & 4.00 & 28.48 & 4.04 & 1.74 & 0.00 \\
    Qwen3.5-9B-SFT & \underline{58.48} & \underline{47.10} & \underline{57.12} & {\bf 54.46} & \underline{51.45} & \underline{41.67} \\
    Qwen3.5-9B-SFT+GRPO & \textbf{60.45} & \textbf{48.40} & \textbf{59.10} & \underline{53.96} & \textbf{51.97} & \textbf{42.18} \\
  \bottomrule
\end{tabular}
\caption{VLM performance on {\task} under  different vision styles or formats. The best results are indicated in bold, while the second-
best are underlined in each group.}
  \label{tab:exp-main-image1}
\end{table*}

\begin{table}
  \centering
  \begin{tabular}{l|cc}
    \toprule
    Model & MathVerse & MMMU \\
    \midrule
    Qwen3.5-9B & 48.94 & 67.52 \\
    Qwen3.5-9B-SFT & 50.76 & 68.05 \\
    Qwen3.5-9B-SFT+GRPO & 51.40 & 68.91 \\
  \bottomrule
\end{tabular}
\caption{Zero-shot generalization performance on out-of-domain multimodal math benchmarks (non-thinking mode).}
\label{tab:exp-mm-math}
\end{table}

\subsection{Generalization to Different Vision Styles}

In this experiment, we verify the model's genuine reasoning ability and demonstrate that its performance remains largely invariant to superficial visual variations.

First, from Table \ref{tab:exp-main-image1}'s results, we can see that our post-trained variants exhibit significantly greater improvements to all surface-level stylistic perturbations than the pre-trained Qwen3.5-9B baseline. This underscores a critical distinction between perceptual invariance—the ability to tolerate changes in color, font, or background—and true reasoning depth. In conclusion, moderate changes in appearance do not substantially alter its reasoning behavior, indicating the of robustness of post-training to nuisance visual factors.

Second, the experiments show that the model perform relatively worse when the box borders are removed. Compared with other style changes, the absence of borders leads to a more noticeable degradation, implying that borders provide an important structural signal for the model. A plausible explanation is that the grid boundaries help the model localize individual cells and identify the spatial positions of the target boxes that require reasoning. Without such explicit delimiters, the model may struggle to segment the visual scene into discrete reasoning units, which in turn harms its ability to track the relationships among cells. We hypothesize that while models tolerate changes in color, font, or background, they still struggle when the visual structure becomes less explicit or when the task demands deeper reasoning across multiple steps.

\subsection{Out-of-Domain Generalization Experiments}

To evaluate the generalization of models trained on \textsc{CrossMath}, we conduct experiments on out-of-domain (OOD) multimodal mathematical reasoning benchmarks: MMMU~\cite{DBLP:conf/cvpr/YueNZ0LZSJRSWYY24} and MathVerse~\cite{DBLP:conf/eccv/ZhangJZLGQZLCQGL24}. In MathVerse, which provides varying levels of textual metadata, we specifically focus on the Vision-Only subset to rigorously isolate the models' visual reasoning capabilities.

As shown in Table~\ref{tab:exp-mm-math}, training on \textsc{CrossMath} consistently yields performance gains. For example, the incremental boost from GRPO over base model ($+2.46\%$ on MathVerse and $+1.39\%$ on MMMU) confirms that reasoning-level optimization is more effective than simple pattern matching for enhancing the model's ability to navigate diverse multi-modal logical challenges. While we do not claim universal transferability across all multimodal tasks, these results demonstrate a significant cross-task synergy within the realm of structured, multi-step mathematical reasoning. Therefore, advancing the visual foundation will likely yield a ``rising tide'' effect, elevating the model's performance across various structured reasoning benchmarks simultaneously.

\subsection{General Discussions}

To further analyze the behavior of other VLMs, we extend our experiments beyond Qwen3.5-9B to include a broader set of models. The results for Qwen3.5-9B are obtained from our local deployment, those for Qwen3.6-Plus are obtained via the OpenRouter API~\footnote{\url{https://openrouter.ai/qwen/qwen3.6-plus}}, and those for the remaining models are collected through Qwen's official API~\footnote{\url{https://modelstudio.console.alibabacloud.com/}}. As shown in Tables~\ref{tab:exp-main-multimodal}, \ref{tab:exp-main-image}, and \ref{tab:exp-main-image1}, several key observations can be drawn:

\noindent {\bf (1) Image-only performance does not scale clearly with model size.} Under the image-only setting, the performance gap among the medium- and large-scale Qwen VLMs is relatively limited and does not show a monotonic scaling trend. For example, Qwen3.5-27B achieves the best micro accuracy, while Qwen3.5-397B-A17B achieves the best macro accuracy. This suggests that simply increasing the size of the language backbone does not consistently improve visual reasoning when the model must rely solely on image inputs. A plausible explanation is that the vision modules across these model variants are largely similar in capacity, making visual perception and visual-to-text grounding, rather than language-model scale, the primary bottleneck in this setting.

\noindent {\bf (2) Textual reasoning exhibits a much clearer scaling trend.} In contrast, under the text-only setting, performance improves substantially as model capacity increases, with Qwen3.5-Plus and Qwen3.5-397B-A17B achieving the strongest results. This pattern is broadly consistent with the scaling behavior commonly observed in LLMs: when the problem is presented in a purely textual form, larger or stronger models are better able to exploit their reasoning capability. Together with the first observation, this result suggests that the main challenge in {\task} is not abstract reasoning itself, but reasoning grounded in visual inputs.

\noindent {\bf (3) Qwen3.6-Plus retains the strongest reasoning capability under multimodal input.} While its text-only performance is comparable to other top-performing models, it achieves the best multimodal results, reaching 90.76 micro accuracy and 79.60 macro accuracy. This suggests that Qwen3.6-Plus is more effective at integrating visual and textual evidence without substantially sacrificing its underlying reasoning capability. In other words, compared with other models whose performance drops more noticeably when visual input is introduced, Qwen3.6-Plus appears to retain a larger proportion of its textual reasoning strength in the multimodal setting.

Overall, these findings suggest that the current limitation of VLMs on {\task} may not primarily stem from insufficient language modeling capacity. Instead, the results point to a modality gap: scaling the language backbone reliably strengthens textual reasoning, but does not yield comparable improvements in image-only reasoning. This further indicates that advancing visual reasoning may require stronger vision modules and more effective cross-modal alignment, rather than relying solely on larger overall model size.

\section{Conclusions}
In this paper, we systematically investigated the reasoning mechanisms of vision-language models (VLMs) to determine whether their success is driven by genuine visual grounding or predominantly relies on their textual backbones. To achieve this, we introduced {\task}, a meticulously curated benchmark that isolates modality-specific reasoning capabilities through strictly identical task-relevant information in text-only, image-only, and image+text formats. Our extensive evaluations demonstrate that state-of-the-art VLMs suffer from a substantial modality gap, predominantly relying on textual shortcuts rather than genuine visual evidence, to the point where visual inputs often act as a distractor. To mitigate this fundamental limitation, we demonstrated that fine-tuning VLMs on a curated {\task} training set significantly enhances performance across all modalities. The consistent improvements observed on external visual reasoning tasks further validate our approach.  Ultimately, we expect this research to drive the creation of VLMs capable of true cross-modal reasoning, empowering future architectures to authentically synergize information across modalities.

\bibliography{nlp}
\bibliographystyle{technical_template}

\appendix

\section*{Appendix}
\section{Example Instruction Templates}

In the appendix, we release the example instruction templates for reference.

\begin{tcolorbox}[
  colback=white, colframe=black, arc=3mm, width=\columnwidth,
  title=\textbf{Example Input for Chain-of-Thought Trajectories Generation},
  coltitle=white, colbacktitle=blue, fonttitle=\bfseries
]
\label{prompt:trajectory}

A **Cross Math Puzzle** is a grid-based arithmetic reasoning task in which numbers and arithmetic operators are arranged in a crossword-like 2D structure. Some cells contain known numbers or symbols (`+', `-', `×', `÷', `='), while some cells are missing and marked as `?'. Valid arithmetic equations are formed along specific **horizontal** and **vertical** paths in the grid. The objective is to infer the missing values so that **all intersecting equations are simultaneously satisfied**.

\vspace{5pt}

Compared with standard single-line arithmetic problems, a cross math puzzle requires the model to:

1. understand the **2D spatial structure** of the puzzle,

2. identify which cells belong to the same equation,

3. reason over **multiple dependent equations**, and

4. maintain consistency across cells shared by both horizontal and vertical equations.

\vspace{5pt}

You are given a **cross math puzzle** in a **textual markdown grid format**.

\vspace{5pt}

Each cell in the grid may contain:

- a number,

- an arithmetic operator (`+', `-', `×', `÷'),

- an equality sign (`='),

- a missing value marked as `?',

- or an empty cell.

\vspace{5pt}

The markdown table represents the **2D spatial layout** of the puzzle. A valid arithmetic equation is formed whenever numbers and operators appear continuously in a horizontal or vertical direction and include an equality sign. The missing cells marked with `?' must be filled with appropriate numbers such that **every horizontal and vertical equation is correct**.

\vspace{5pt}

You are required to reason as **concisely** as you can, while keeping compulsory reasoning steps.

\vspace{5pt}
Output Instructions:

Thinking step-by-step and output the intermediate reasoning steps.

Then provide the final answer enclosed within `$<$answer$>$' and `$<$/answer$>$' tags.

- List the filled blanks in **left-to-right, top-to-bottom** order.

- Include **only** the blanks that you fill.

- Separate each number or operator with a **single space**.

- Do **not** include any additional explanation, punctuation, or formatting outside the answer tags.

\vspace{5pt}

Here are the cross math puzzle:

\vspace{5pt}

$|$\;\;\;\;\;$|$\;\;\;\;\;$|$\;\;\;\;\;$|$\;\;\;\;\;$|$\;?\;\;$|$\;\;\;\;\;$|$\;\;\;\;\;$|$

$|$\;\;\;\;\;$|$\;\;\;\;\;$|$\;\;\;\;\;$|$\;\;\;\;\;$|$\;+\;\;$|$\;\;\;\;\;$|$\;\;\;\;\;$|$

$|$\;28\;$|$\;\;\;\;\;$|$\;?\;\;$|$\;÷\;\;$|$\;3\;\;$|$\;=\;\;$|$\;31\;$|$

$|$\;+\;\;$|$\;\;\;\;\;$|$\;\;\;\;\;$|$\;\;\;\;\;$|$\;=\;\;$|$\;\;\;\;\;$|$\;\;\;\;\;$|$

$|$\;?\;\;$|$\;÷\;\;$|$\;5\;\;$|$\;=\;\;$|$\;9\;\;$|$\;\;\;\;\;$|$\;\;\;\;\;$|$

$|$\;=\;\;$|$\;\;\;\;\;$|$\;×\;\;$|$\;\;\;\;\;$|$\;\;\;\;\;$|$\;\;\;\;\;$|$\;\;\;\;\;$|$

$|$\;73\;$|$\;\;\;\;\;$|$\;?\;\;$|$\;+\;\;$|$\;57\;$|$\;=\;\;$|$\;65\;$|$

$|$\;\;\;\;\;$|$\;\;\;\;\;$|$\;=\;\;$|$\;\;\;\;\;$|$\;\;\;\;\;$|$\;\;\;\;\;$|$\;\;\;\;\;$|$

$|$\;\;\;\;\;$|$\;\;\;\;\;$|$\;40\;$|$\;\;\;\;\;$|$\;\;\;\;\;$|$\;\;\;\;\;$|$\;\;\;\;\;$|$

\end{tcolorbox}

\begin{tcolorbox}[
  colback=white, colframe=black, arc=3mm, width=\columnwidth,
  title=\textbf{Example Input for Image-Only Query},
  coltitle=white, colbacktitle=red, fonttitle=\bfseries
]

A **Cross Math Puzzle** is a grid-based arithmetic reasoning task in which numbers and arithmetic operators are arranged in a crossword-like 2D structure. Some cells contain known numbers or symbols (`+', `-', `×', `÷', `='), while some cells are missing and marked as `?'. Valid arithmetic equations are formed along specific **horizontal** and **vertical** paths in the grid. The objective is to infer the missing values so that **all intersecting equations are simultaneously satisfied**.

\vspace{5pt}

Compared with standard single-line arithmetic problems, a cross math puzzle requires the model to:

1. understand the **2D spatial structure** of the puzzle,

2. identify which cells belong to the same equation,

3. reason over **multiple dependent equations**, and

4. maintain consistency across cells shared by both horizontal and vertical equations.

\vspace{5pt}

You are given a **cross math puzzle** in a **image format**.

\vspace{5pt}

The puzzle is displayed as a 2D grid in which some cells contain numbers and arithmetic symbols (`+', `-', `×', `÷', `='), while other cells are blank or unknown. Unknown cells correspond to values that must be inferred.

\vspace{5pt}

The model must first understand the **visual layout** of the puzzle: it needs to recognize the cell positions, read the symbols and numbers, determine which cells form horizontal and vertical equations, and then solve for the missing values so that **all equations in the grid are satisfied simultaneously**.

\vspace{5pt}
Output Instructions:

Thinking step-by-step and output the intermediate reasoning steps.

Then provide the final answer enclosed within `$<$answer$>$' and `$<$/answer$>$' tags.

- List the filled blanks in **left-to-right, top-to-bottom** order.

- Include **only** the blanks that you fill.

- Separate each number or operator with a **single space**.

- Do **not** include any additional explanation, punctuation, or formatting outside the answer tags.

\vspace{5pt}

Here are the cross math puzzle:

\vspace{5pt}

\includegraphics[width=0.25\linewidth]{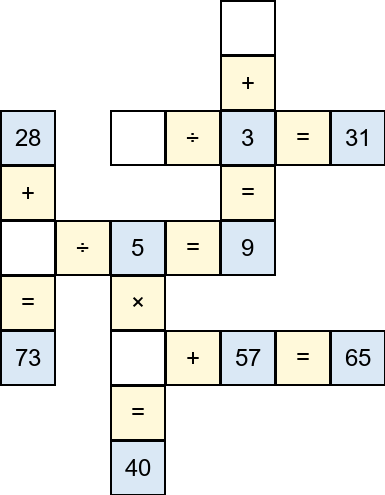}

\end{tcolorbox}

\begin{tcolorbox}[
  colback=white, colframe=black, arc=3mm, width=\columnwidth,
  title=\textbf{Example Input for Image + Text Query},
  coltitle=white, colbacktitle=gray, fonttitle=\bfseries
]

A **Cross Math Puzzle** is a grid-based arithmetic reasoning task in which numbers and arithmetic operators are arranged in a crossword-like 2D structure. Some cells contain known numbers or symbols (`+', `-', `×', `÷', `='), while some cells are missing and marked as `?'. Valid arithmetic equations are formed along specific **horizontal** and **vertical** paths in the grid. The objective is to infer the missing values so that **all intersecting equations are simultaneously satisfied**.

\vspace{5pt}

Compared with standard single-line arithmetic problems, a cross math puzzle requires the model to:

1. understand the **2D spatial structure** of the puzzle,

2. identify which cells belong to the same equation,

3. reason over **multiple dependent equations**, and

4. maintain consistency across cells shared by both horizontal and vertical equations.

\vspace{5pt}

You are given a **cross math puzzle** in **both image format and textual markdown format**.

\vspace{5pt}

The image provides the visual structure and appearance of the puzzle, while the markdown table provides an explicit symbolic representation of the same grid.

\vspace{5pt}

The model may use both modalities together:

- the **image** helps recover spatial arrangement and resolve layout ambiguity,

- the **textual grid** helps read symbols and numbers more reliably,

- combining both sources improves robustness in identifying equations and solving the puzzle.

\vspace{5pt}

The goal is to infer the missing values marked by `?' so that **all horizontal and vertical arithmetic equations are simultaneously satisfied**.

\vspace{5pt}
Output Instructions:

Thinking step-by-step and output the intermediate reasoning steps.

Then provide the final answer enclosed within `$<$answer$>$' and `$<$/answer$>$' tags.

- List the filled blanks in **left-to-right, top-to-bottom** order.

- Include **only** the blanks that you fill.

- Separate each number or operator with a **single space**.

- Do **not** include any additional explanation, punctuation, or formatting outside the answer tags.

\vspace{5pt}

Here are the cross math puzzle:

\vspace{5pt}

$|$\;\;\;\;\;$|$\;\;\;\;\;$|$\;\;\;\;\;$|$\;\;\;\;\;$|$\;?\;\;$|$\;\;\;\;\;$|$\;\;\;\;\;$|$

$|$\;\;\;\;\;$|$\;\;\;\;\;$|$\;\;\;\;\;$|$\;\;\;\;\;$|$\;+\;\;$|$\;\;\;\;\;$|$\;\;\;\;\;$|$

$|$\;28\;$|$\;\;\;\;\;$|$\;?\;\;$|$\;÷\;\;$|$\;3\;\;$|$\;=\;\;$|$\;31\;$|$

$|$\;+\;\;$|$\;\;\;\;\;$|$\;\;\;\;\;$|$\;\;\;\;\;$|$\;=\;\;$|$\;\;\;\;\;$|$\;\;\;\;\;$|$

$|$\;?\;\;$|$\;÷\;\;$|$\;5\;\;$|$\;=\;\;$|$\;9\;\;$|$\;\;\;\;\;$|$\;\;\;\;\;$|$

$|$\;=\;\;$|$\;\;\;\;\;$|$\;×\;\;$|$\;\;\;\;\;$|$\;\;\;\;\;$|$\;\;\;\;\;$|$\;\;\;\;\;$|$

$|$\;73\;$|$\;\;\;\;\;$|$\;?\;\;$|$\;+\;\;$|$\;57\;$|$\;=\;\;$|$\;65\;$|$

$|$\;\;\;\;\;$|$\;\;\;\;\;$|$\;=\;\;$|$\;\;\;\;\;$|$\;\;\;\;\;$|$\;\;\;\;\;$|$\;\;\;\;\;$|$

$|$\;\;\;\;\;$|$\;\;\;\;\;$|$\;40\;$|$\;\;\;\;\;$|$\;\;\;\;\;$|$\;\;\;\;\;$|$\;\;\;\;\;$|$

\vspace{5pt}

\includegraphics[width=0.25\linewidth]{res/math_puzzle_0003_blank.png}

\end{tcolorbox}

\begin{tcolorbox}[
  colback=white, colframe=black, arc=3mm, width=\columnwidth,
  title=\textbf{Example Input for Text-Only Query},
  coltitle=white, colbacktitle=blue, fonttitle=\bfseries
]

A **Cross Math Puzzle** is a grid-based arithmetic reasoning task in which numbers and arithmetic operators are arranged in a crossword-like 2D structure. Some cells contain known numbers or symbols (`+', `-', `×', `÷', `='), while some cells are missing and marked as `?'. Valid arithmetic equations are formed along specific **horizontal** and **vertical** paths in the grid. The objective is to infer the missing values so that **all intersecting equations are simultaneously satisfied**.

\vspace{5pt}

Compared with standard single-line arithmetic problems, a cross math puzzle requires the model to:

1. understand the **2D spatial structure** of the puzzle,

2. identify which cells belong to the same equation,

3. reason over **multiple dependent equations**, and

4. maintain consistency across cells shared by both horizontal and vertical equations.

\vspace{5pt}

You are given a **cross math puzzle** in a **textual markdown grid format**.

\vspace{5pt}

Each cell in the grid may contain:

- a number,

- an arithmetic operator (`+', `-', `×', `÷'),

- an equality sign (`='),

- a missing value marked as `?',

- or an empty cell.

\vspace{5pt}

The markdown table represents the **2D spatial layout** of the puzzle. A valid arithmetic equation is formed whenever numbers and operators appear continuously in a horizontal or vertical direction and include an equality sign. The missing cells marked with `?' must be filled with appropriate numbers such that **every horizontal and vertical equation is correct**.

\vspace{5pt}
Output Instructions:

Thinking step-by-step and output the intermediate reasoning steps.

Then provide the final answer enclosed within `$<$answer$>$' and `$<$/answer$>$' tags.

- List the filled blanks in **left-to-right, top-to-bottom** order.

- Include **only** the blanks that you fill.

- Separate each number or operator with a **single space**.

- Do **not** include any additional explanation, punctuation, or formatting outside the answer tags.

\vspace{5pt}

Here are the cross math puzzle:

\vspace{5pt}

$|$\;\;\;\;\;$|$\;\;\;\;\;$|$\;\;\;\;\;$|$\;\;\;\;\;$|$\;?\;\;$|$\;\;\;\;\;$|$\;\;\;\;\;$|$

$|$\;\;\;\;\;$|$\;\;\;\;\;$|$\;\;\;\;\;$|$\;\;\;\;\;$|$\;+\;\;$|$\;\;\;\;\;$|$\;\;\;\;\;$|$

$|$\;28\;$|$\;\;\;\;\;$|$\;?\;\;$|$\;÷\;\;$|$\;3\;\;$|$\;=\;\;$|$\;31\;$|$

$|$\;+\;\;$|$\;\;\;\;\;$|$\;\;\;\;\;$|$\;\;\;\;\;$|$\;=\;\;$|$\;\;\;\;\;$|$\;\;\;\;\;$|$

$|$\;?\;\;$|$\;÷\;\;$|$\;5\;\;$|$\;=\;\;$|$\;9\;\;$|$\;\;\;\;\;$|$\;\;\;\;\;$|$

$|$\;=\;\;$|$\;\;\;\;\;$|$\;×\;\;$|$\;\;\;\;\;$|$\;\;\;\;\;$|$\;\;\;\;\;$|$\;\;\;\;\;$|$

$|$\;73\;$|$\;\;\;\;\;$|$\;?\;\;$|$\;+\;\;$|$\;57\;$|$\;=\;\;$|$\;65\;$|$

$|$\;\;\;\;\;$|$\;\;\;\;\;$|$\;=\;\;$|$\;\;\;\;\;$|$\;\;\;\;\;$|$\;\;\;\;\;$|$\;\;\;\;\;$|$

$|$\;\;\;\;\;$|$\;\;\;\;\;$|$\;40\;$|$\;\;\;\;\;$|$\;\;\;\;\;$|$\;\;\;\;\;$|$\;\;\;\;\;$|$

\end{tcolorbox}

\end{document}